\documentclass[a4paper,fleqn]{cas-dc}

\usepackage[numbers]{natbib}
\usepackage{graphicx}
\graphicspath{{figures/}}
\usepackage{array}
\usepackage{booktabs}
\usepackage{caption}
\usepackage{subcaption}
\usepackage{amsmath,amssymb}

\usepackage{pifont}
\newcommand{\cmark}{\ding{51}}
\newcommand{\xmark}{\ding{55}}

\usepackage[linesnumbered,ruled,vlined]{algorithm2e}

\SetKwInput{KwInput}{Input}  
\SetKwInput{KwOutput}{Output}

\begin{document}

\let\WriteBookmarks\relax
\def\floatpagepagefraction{1}
\def\textpagefraction{.001}

\shorttitle{Designing ECG Monitoring Healthcare System with Federated Transfer Learning and Explainable AI}
\shortauthors{Ali Raza et~al.}

\title [mode = title]{Designing ECG Monitoring Healthcare System with Federated Transfer Learning and Explainable AI}

\author[1,2]{Ali~Raza}[orcid=0000-0001-8326-8325]
\cormark[1]
\author[1]{Kim~Phuc~Tran}
\author[1]{Ludovic~Koehl}[orcid=0000-0002-3404-8462]
\author[2]{Shujun~Li}[orcid=0000-0001-5628-7328]

\address[1]{University of Lille, ENSAIT, GEMTEX–Laboratoire de Génie et Matériaux Textiles, F-59000 Lille, France}

\address[2]{School of Computing \& Institute of Cyber Security for Society (iCSS), University of Kent, UK}

\cortext[cor1]{Corresponding author}

\nonumnote{The authors can be contacted via ali.raza@ensait.fr or ar718@kent.ac.uk (Ali Raza), kim-phuc.tran@ensait.fr (Kim Phuc Tran), ludovic.koehl@ensait.fr (Ludovic Koehl), hooklee@gmail.com or S.J.Li@kent.ac.uk (Shujun Li).}

\begin{abstract}
Deep learning plays a vital role in classifying different arrhythmias using electrocardiography (ECG) data. Nevertheless, training deep learning models normally requires a large amount of data and can lead to privacy concerns. Unfortunately, a large amount of healthcare data cannot be easily collected from a single silo. Additionally, deep learning models are like black-box, with no explainability of the predicted results, which is often required in clinical healthcare. This limits the application of deep learning in real-world health systems.

In this paper, to address the above-mentioned challenges, we design a novel end-to-end framework in a federated setting for ECG-based healthcare using explainable artificial intelligence (XAI) and deep convolutional neural networks (CNN). The federated setting is used to solve challenges such as data availability and privacy concerns. Furthermore, the proposed framework effectively classifies different arrhythmias using an autoencoder and a classifier, both based on a CNN. Additionally, we propose an XAI-based module on top of the proposed classifier for interpretability of the classification results, which helps clinical practitioners to interpret the predictions of the classifier and to make quick and reliable decisions. The proposed framework was trained and tested using the baseline Massachusetts Institute of Technology - Boston's Beth Israel Hospital (MIT-BIH) Arrhythmia database. The trained classifier outperformed existing work by achieving accuracy up to 94.5\% and 98.9\% for arrhythmia detection using noisy and clean data, respectively, with five-fold cross-validation. We also propose a new communication cost reduction method to reduce the communication costs and to enhance the privacy of users' data in the federated setting. While the proposed framework was tested and validated for ECG classification, it is general enough to be extended to many other healthcare applications.
\end{abstract}


\maketitle

\section{Introduction}

With the increase of internet of things (IoT) devices being used in the 21st century, massive amount of data has been generated \cite{marjani2017big}. IoT devices are capable of collecting an enormous amount of data each day \cite{mourtzis2016industrial}. This collection of data and the exponentially increasing computational resources have unlocked new dimensions in the information technology sector, especially in deep learning (DL) \cite{mohammadi2018deep}. Although deep learning is a quite old concept \cite{shavlik1990readings} but owing to limited data and computational resources available in the past its use was limited. However, thanks to the internet, IoT devices and the increasing computational power, nowadays we can see deep learning revolutionizing nearly every field, including healthcare \cite{esteva2019guide}, economics \cite{ozbayoglu2020deep}, manufacturing \cite{wang2018deep}, agriculture \cite{kamilaris2018deep}, and military \cite{hossain2019comprehensive}.

In regards to healthcare applications, a lot of data is being generated across the globe and it has quite unique properties. Most of the data related to healthcare are multi-dimensional, this makes the use of classical machine learning (ML) models, for example, decision trees and random forests, quiet challenging and complex. However, the new generation machine learning models, especially the deep learning based ones, can solve issues related to multi-dimensional data due to its capability of self learning \cite{georgiou2020survey}. In the healthcare industry, deep learning has played a critical role, e.g., to help diagnose  life threatening diseases \cite{miotto2018deep}. Nevertheless, it has some limitations \cite{kumar2017deep}. First, to train a deep learning model a large amount of training data is needed, but each silo (for example, a hospital) can have a very limited amount of data, so a single source of data can be insufficient to train a good deep learning model. A solution to this is to collect data from multiple sources and then train the model on the collected data. One major issue of this approach is about privacy concerns \cite{ji2014differential}. As medical data are highly sensitive and private data, some individual sources may not be willing to share their data with a central data collector \cite{van2016privacy}.

In 2016 Google came forward with an idea called federated learning to solve the conflict between data availability and privacy concerns \cite{konevcny2016federated}. The basic idea behind federated learning is to collaboratively train a machine learning model without centralized training data. Federated learning enables edge devices or servers  with sufficient computational power (e.g., home computers, mobile phones, wearables and other IoT devices) to collaboratively learn a shared machine learning model while keeping all the training data on local devices, decoupling the ability to do machine learning from the need to store the data centrally at a single server or in the cloud. Although deep learning with a federated setting can solve the issues mentioned earlier, there exists the problem of explainability in deep learning. Since the deep learning models are generally black box models, with no reasonable explanation for a given prediction. This ambiguity causes a limitation of deep learning in healthcare, because a clinical practitioner should know the reason for a prediction by a deep learning model \cite{gunning2017explainable}. To address the problem of explainability in deep learning models, researchers have proposed different solutions \cite{samek2017explainable, choo2018visual,mousavi2020han}. For instance, Selvaraju proposed a method called  Gradient-weighted Class Activation Mapping (Grad-CAM) \cite{selvaraju2016grad} to visualize input regions that are important for predictions. From such values, we can have an idea about where exactly the machine learning model is focusing while making a prediction and thus the reason. Explainability is important in healthcare, because to convince a clinical healthcare practitioner and a patient we need to give them the reason behind a certain prediction for sample input.

In regards to the application of deep learning in healthcare, electrocardiogram (ECG) classification is a very important routine task. Many machine learning based solutions have been proposed for analyzing and classifying ECG data \cite{mousavi2020han,pyakillya2017deep, mathews2018novel, murat2020application,chen2020fedhealth,oh2018automated, liaqat2020detection,erdenebayar2019automatic,atal2020arrhythmia,acharya2017automated,yao2017atrial,nurmaini2020robust}. However, most of these works are based on a centralized machine learning architecture, thereafter they are prone to issues like privacy concerns and data availability. Moreover, since most of the real-time ECG data is noisy, they cannot perform well in real time because they are being trained on preprocessed (cleaner) data. Furthermore, they do not provide explainability/interpretability (we use explainability and interpretability interchangeably through out this article  ), which is one of the key requirements in deep learning based clinical healthcare. Hence, this limits their real-time application. To address all of the above-mentioned challenges, in this paper, we propose an end-to-end explainable healthcare framework in a federated setting. The proposed Framework consists of three main parts: an autoencoder, a classifier and an XAI module. Firstly, we propose a novel deep convolutional neural network (CNN) based autoencoder, which is used to denoise the raw ECG signals from the subject directly. Secondly, we propose a novel CNN-based classifier, which uses transfer learning to classify the raw time series of ECG data. Thirdly, we adopt the Grad-CAM model \cite{selvaraju2016grad} in the framework to explain the classification results in a novel and reliable pattern. Additionally, we propose a custom communication cost reduction approach that  reduces the communication cost and increases the privacy protection of the framework.
\subsection{Contributions}
The main contributions of this paper are as follows:
\begin{enumerate}
\item We propose an end-to-end framework which is the first federated transfer learning and explainable-AI based framework for healthcare. It aggregates the data from different edge devices (hospitals, users) without compromising privacy and security, provides relatively personalized model learning through knowledge transfer and provides interpretability of the results, which is one of the key requirements in applications like healthcare. In addition to interpretability, the proposed XAI module can be used to recognize new potential patterns leading to trigger heart arrhythmias.

\item We propose a novel 1-Dimensional CNN-based autoencoder in a federated setting to efficiently denoise the raw time series of ECG signal collected data from patients. The autoencoder provides a denoised version of the input, which we use for further classification and explanation of the predictions. 
	
\item With the help of transfer learning, we use the encoder part of the proposed autoencoder to make a novel 1-Dimensional CNN-based classifier to classify given ECG data into five classes: non-ecotic beats (N), supraventricular ectopic beats (S), ventricular ectopic beats (V), fusion Beats (F), and unknown beats (Q).
	
\item We propose a novel module, called XAI module for interpretability of predictions of the proposed classifier. The proposed XAI module is combined with the proposed classifier to interpret and explain the decision making process of the classifier. The XAI module can be used with every updated classier locally at the edge devices in the federated setting, and it does not need any pre-training. 
\item We propose a new communication cost reduction method for the federated learning in the proposed framework, which not only reduces the communication costs but also increases the privacy of the classical federated learning method. Furthermore, the proposed method can be integrated into existing cost optimization algorithms to enhance their cost effectiveness and privacy protection level. 
	
\item We used the MIT-BIH Arrhythmia Database \cite{moody2001impact} to train our proposed framework. It is important to note that to make the data more realistic, we first upsample the data to create more data samples, and then add 10-30\% random noise. The proposed framework shows excellent performance by providing an overall accuracy of $94.5\%$ using noisy data and overall accuracy of $98.9\%$ on the clean data in the original MIT-BIH database. Moreover, we evaluated the performance of the proposed framework using four standard metrics: classification accuracy, precision, recall and F1-score.
	
\item The proposed framework additionally boasts desirable features: interpretability of the results by using the proposed XAI module, and efficient classification of the ECG. Additionally, it provides an enhanced level of privacy protection to users because of the federated setting and the proposed communication cost reduction method.
\end{enumerate}

The rest of the paper is organized as follows. Section~\ref{sec:related_work} presents related work and background. Section~\ref{sec:Proposed_Method} discusses  detailed description of the proposed framework. Sections~\ref{sec:experiments} and \ref{sec:Analysis} present the experimental setup and performance evaluation, respectively. Section~\ref{sec:conclusion} concludes the article.

\section{Related Work and Background}
\label{sec:related_work}
 Intelligent systems have helped us achieve efficient solutions. Various types of intelligent systems, such as,  intelligent systems for modeling uncertainties by robust optimization \cite{yang2021robust}, intelligent parking lots for electrical vehicles \cite{liu2020igdt}, and other wide ranges of applications \cite{angelov2010evolving} have been a focus of academia and industries. In recent decades, intelligent systems based on ML have been studied in a wide range of applications. For example, its applications have been studied for cyber security \cite{handa2019machine}, economics and agriculture \cite{storm2020machine}, and in healthcare \cite{shailaja2018machine}. The use of machine learning in healthcare has been widely studied ranging from detection and diagnosis of different diseases, such as, melanoma \cite{razmjooy2018hybrid,parsian2017hybrid} and cancer \cite{xu2020computer,kourou2015machine}. Owing to the importance of machine learning in healthcare-related applications, in this section we review the literature of machine learning in  healthcare, with a special focus on machine learning for ECG analysis. \color{black}
\subsection{Machine Learning in Healthcare}

Certain activities in our body are governed by signals of some cognitive diseases \cite{atkinson2007cognitive}. For example, a changing gait may result from a stroke. A number of researchers proposed to monitor users' activities using wearable sensors, with the help of which different human body activities can be recognized \cite{chen2019cross, mukhopadhyay2014wearable, lara2012survey}. Based on monitoring of such activities, early prognosis of health issues can be identified. In this regards, there has been significant development in the utilization of ML and DL technologies in healthcare. While such technologies will probably never completely replace clinical practitioners, they can transform the healthcare sector, benefiting both patients and providers \cite{miotto2018deep, bhardwaj2017study, manogaran2017survey, fakoor2013using}.

In regards to ECG analysis in healthcare, ML and DL play a vital role. Researchers have proposed many methods for ECG classification into arrhythmia types \cite{jambukia2015classification, roopa2017survey, sahoo2020machine, atal2020arrhythmia, liaqat2020detection}. Rubin et al.~\cite{rubin2017recognizing} applied deep learning to the task of automated cardiac auscultation, i.e., recognizing abnormalities in heart sounds. They described an automated heart sound classification algorithm that combines the use of time-frequency heat map representations with a deep CNN. Their CNN architecture is trained using a modified loss function that directly optimizes the trade-off between sensitivity and specificity. Gjoreski et al.~\cite{gjoreski2020machine} presented a method for chronic heart failure (CHF) detection based on heart sounds. The method combines classic ML and end-to-end DL models. The classic ML model learns from expert features, and the DL model learns from a spectro-temporal representation of the signal. Moreover, in order to enable intelligent classification of arrhythmias with high accuracy, Huang et al.~\cite{huang2020accurate} presented an intelligent ECG classifier using the fast compression residual convolutional neural networks (FCResNet).

Although the aforementioned work seems promising, they may find limited applicability in real world because they use centralized data collection techniques. As discussed earlier that it may cause privacy concerns among users and data owners. Thereafter, traditional centralized healthcare applications find limited applicability due to privacy concerns \cite{inkster2018china, liu2021machine, waheed2020security}. To address the privacy issues in machine learning, researchers have been working on Federated learning (FL) and Transfer learning (TF). Federated learning (FL) was introduced by Google \cite{konevcny2016federated}. The key idea is to train ML models with privacy by design at the architectural level. FL trains a machine learning model in a distributed architecture, where the edged devices train their own ML model on their local data and a central global server aggregates all of the locally trained models and distribute the aggregated model back to all nodes on the network (more details about FL can be found in Section~\ref{subsec:federated}). Due to its privacy preserving and efficient communication constraints, FL finds a number of applications in healthcare \cite{yang2019federated}. Xu et al.~\cite{xu2021federated} summarized the general solutions to the statistical challenges, system challenges, and privacy, and point out the implications and potentials of FL's application in healthcare. They show that training the model in the federated learning framework leads to comparable performance to the traditional centralized learning setting. Transfer learning (TF) aims at transferring knowledge from an existing trained model to a new model. The key idea is to reduce the distribution divergence between different models. To this end, there are mainly two general approaches: instance reweighting \cite{huang2012boosting} and feature matching \cite{qin2019cross}. Recently, deep transfer learning methods have made considerable success in many application fields. Chen et al.~\cite{chen2020fedhealth} proposed FedHealth, the first federated transfer learning framework for wearable healthcare to tackle privacy and security challenges. FedHealth performs data aggregation through federated learning, and then builds relatively personalized models by transfer learning. FedHealth makes it possible to do deep transfer learning in the federated learning framework without accessing the raw user data. However, there are certain limitations to it. Firstly, it does not provide the explainability of the predictions, which is often required in sensitive domains like healthcare. Secondly, it does not accommodate any mechanism to denoise the raw signals, which often contain random noise and dealing with the random noise is quite challenging.

In other words, regarding the application of ML and DL healthcare, a lot of promising work has been done as discussed above. However, some of those works are vulnerable to privacy issues. Research work like FedHealth tries to address the issues of privacy concerns using FL and TL architecture. Nevertheless, works like FedHealth have the limitation of explainability, as discussed earlier. Thus there is a need for research work to address such challenges.

\subsection{Autoencoder}

Autoencoder \cite{zhai2018autoencoder} is an unsupervised neural network that learns the best encoding-decoding scheme from data. In general, it consists of an input layer, an output layer, an encoder neural network, a decoder neural network, and a latent space. When the data is fed to the network, the encoder compresses data into a latent space, whereas the decoder decompresses the encoded representation into the output layer. The encoded-decoded output is then compared with the initial data and the error is backpropagated through the architecture to update the weights of the network \cite{nguyen2021forecasting}. Given the input $x\in R^{m}$, the encoder compresses $x$ to obtain an encoded representation $z=e(x)\in R^{n}$. The decoder reconstructs this representation to give the output $x^{'}=d(z)\in R^{m}$. The autoencoder is trained by minimizing the reconstruction error $L$, defined by the following equation:
\begin{equation}
L=\frac{1}{n}\sum_{i=1}^{n}(Y_{i}-Y_{i}^{'})^{2},
\end{equation}
where $Y_{i}$ is the true label, $Y_{i}^{'}$ is the predicted label, and $n$ is the total number of samples. An ideal autoencoder simply copies the input to the output, whereas keeping the latent space to have a smaller dimension than the input. The autoencoder learns the most salient features of the training data, i.e., it reduces the data dimensions while keeping the important information of the data.

Since being proposed, many researchers have proposed many optimized approaches of autoencoder, such as sparse autoencoder, denoising autoencoder, contractive autoencoder, and convolutional autoencoder \cite{jia2018survey}. We can achieve two main tasks from autoencoders: denoising and dimensionality reduction. In this study, we build a denoising autoencoder, which is an extension of simple autoencoders. It is worth noting that denoising autoencoders were not originally meant to automatically denoise an input. Instead, the denoising autoencoder procedure was invented to help:
\begin{enumerate}
\item the hidden layers of the autoencoder learn more robust filters,
	
\item reduce the risk of overfitting in the autoencoder, and
	
\item prevent the autoencoder from learning a simple identity function.
\end{enumerate}
In denoising autoencoders noise is  stochastically (i.e., randomly) added to the input data, and then the autoencoder is trained to recover the original, non-perturbed signal.

\subsection{Federated Learning}
\label{subsec:federated}

\begin{figure}[!ht]
\centering
\includegraphics[width=\linewidth]{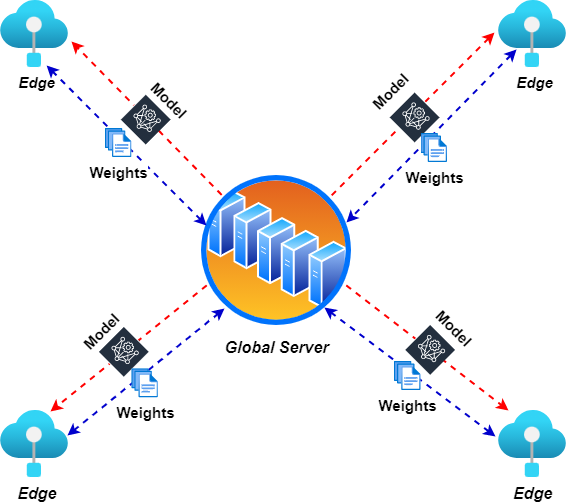}
\caption{Architecture of Federated Learning}
\label{fig:fl-architecture}
\end{figure}

Federated machine (FL) learning was first proposed by Google \cite{konevcny2016federated}, an overview of FL is shown in Figure~\ref{fig:fl-architecture}. In FL settings machine learning models are trained based on distributed edge devices all over the world. The key idea is to protect user data during the process. FL has the ability to resolve the data islanding problems by privacy-preserving model training in the network.

It works like this: an edge (client) device downloads the current model, improves it by learning from data on its local data, and then summarizes the changes as a small focused update. Only this update to the model is sent to the cloud, using encrypted communication, where it is aggregated with other user updates to improve the global shared model. All the training data remains on local devices, and no individual updates are stored in the cloud. Federated Learning allows for smarter models, lower latency, and less power consumption, while ensuring privacy. This approach has another benefit: in addition to providing an update to the global shared model, the improved model on the local edge device can also be used immediately, powering experiences personalized by the use of IoT devices.

\subsection{Transfer Learning}

Transfer learning aims at transferring knowledge from existing domains to a new domain. The key idea is to reduce the distribution divergence between different domains. Here are mainly two types of transfer learning: instance
reweighting \cite{huang2012boosting} and feature matching \cite{qin2019cross}. Recently, deep transfer learning methods have made considerable success in many application fields.

\subsection{Explainable Artificial Intelligence}

Explainable Artificial Intelligence (XAI) \cite{gunning2017explainable} lets humans understand and articulate how an AI system made a decision. XAI is a set of processes and methods that allows human users to comprehend and trust the results and output created by machine learning algorithms. XAI is used to describe an AI model, its expected impact and potential biases. It helps characterize model accuracy, fairness, transparency and outcomes in AI-powered decision making. XAI is crucial for an organization in building trust and confidence when putting AI models into production. AI explainability also helps an organization adopt a responsible approach to AI development. There are many advantages to understanding how an AI-enabled system has led to a specific output. Explainability can help developers ensure that the system is working as expected, it might be necessary to meet regulatory standards, or it might be important in allowing those affected by a decision to challenge or change that outcome. Recent research suggest that it will be of key importance in marketing \cite{yilmazer2021shelf}, healthcare, manufacturing, insurance, and automobiles \cite{dovsilovic2018explainable}.

\section{The Proposed Framework}
\label{sec:Proposed_Method}

Before describing our proposed framework in detail, let us explain the research problem first. Given data on $N$ different edge nodes (since we are using cross-silo federated learning, each  edge node can represent a different organization, i.e., hospital) represented by $E=\{E_{1}, E_{2}, \ldots, E_{N}\}$ and the data of each $E_{i}$ (here $i= 1, 2, \ldots, N$) is given by $\{ D_{1},D_{2},\ldots,D_{i}\}$, respectively. A conventional machine learning model, denoted by ConMOD, can be trained by combining all the data $D=\{D_{1}, D_{2}, \ldots, D_{i}\}$. The data from different edge nodes have different distributions. However, in our problem, we want to collaborate all the data to train a federated transfer learning model, denoted by FedMOD, where any user $E_{i}$ does not expose its data $D_{i}$ to others. Assume that AccFed represents the accuracy of FedMOD and $\text{AccCon}_i$ represents the accuracy of each locally trained model of $E_{i}$, then one of the objectives of our proposed method is to ensure that the accuracy of AccFed is close to or superior to each $\text{AccCon}_i$.

\begin{figure*}[!ht]
\centering
\includegraphics[width=0.9\linewidth]{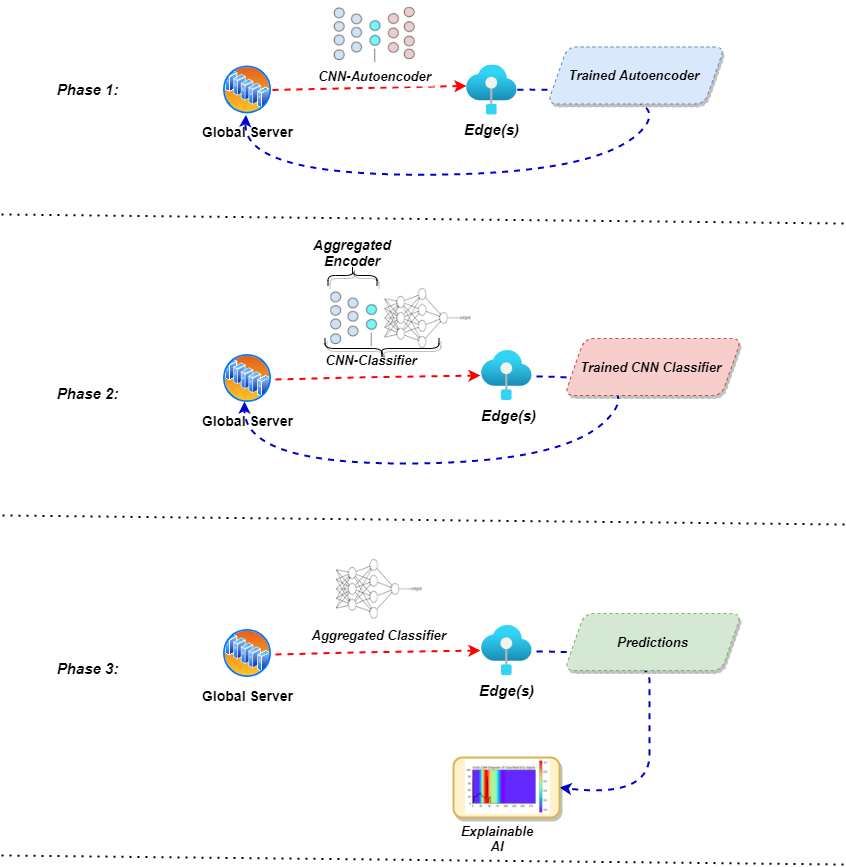}
\caption{An overview of the proposed framework}
\label{fig:framework_overview}
\end{figure*}

The proposed framework aims to achieve accurate and efficient personal healthcare through federated transfer learning and XAI without compromising privacy. Figure~\ref{fig:framework_overview} gives an overview of the proposed method. The proposed method consists of three major parts, the autoencoder, the classifier and the XAI module, which are discussed below in the following three sub-sections. The final sub-section \ref{subsec:Learning_Process} discusses the learning process.

\subsection{CNN-based Autoencoder}

\begin{figure*}[!ht]
\centering
\includegraphics[width=0.9\linewidth]{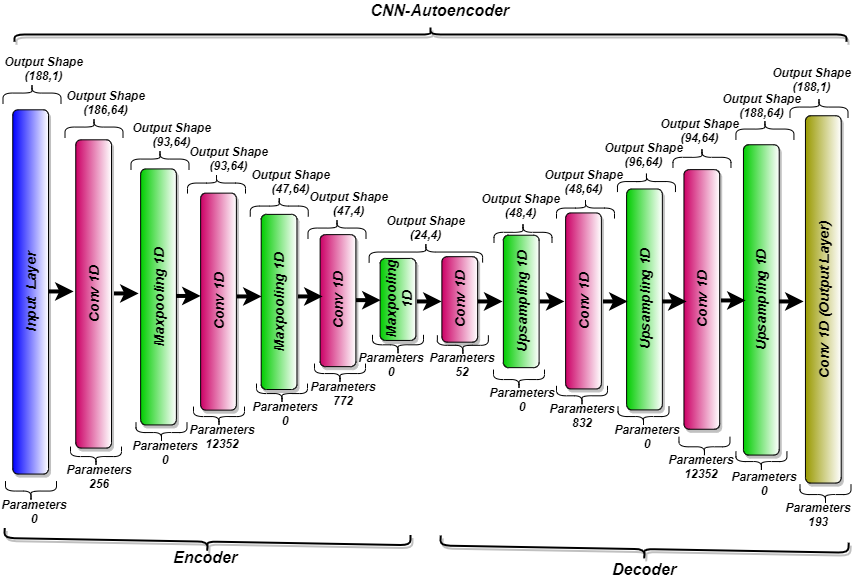}
\caption{The architecture of the proposed denoising autoencoder}
\label{fig:autoencoder}
\end{figure*}

In order to denoise the raw input signal from ECG devices, we proposed an autoencoder. The proposed autoencoder is shown in Figure~\ref{fig:autoencoder}. It consists of an input layer, an output layer and 12 hidden layers. Among the hidden layers, there are 6 convolutional layers, 3 maxpooling layers and 3 upsampling layers. Furthermore, the CNN-autoencoder is virtually divided into two parts: Encoder and Decoder. The encoder consists of the input layer, 3 maxpooling layers and 3 convolutional layers in an alternate fashion. On other hand, the decoder consists of 3 upsampling layers, 3 convolutional layers and a convolutional output layer. In the proposed autoencoder, we use a varying learning rate to keep the training process efficient while keeping the reconstruction loss $L$ as small as possible. Equation~\eqref{eq:lr} gives the mathematical representation of the learning rate (lr) used.
\begin{equation}\label{eq:lr}
\text{lr}=
\begin{cases}
	0.01, & \text{if epoch $\le$ 40},\\
	\text{lr}\times e^{-0.1}, & \text{otherwise}.
\end{cases}
\end{equation}

\subsection{CNN-based Classifier}

\begin{figure*}[!ht]
\centering
\includegraphics[width=0.9\linewidth]{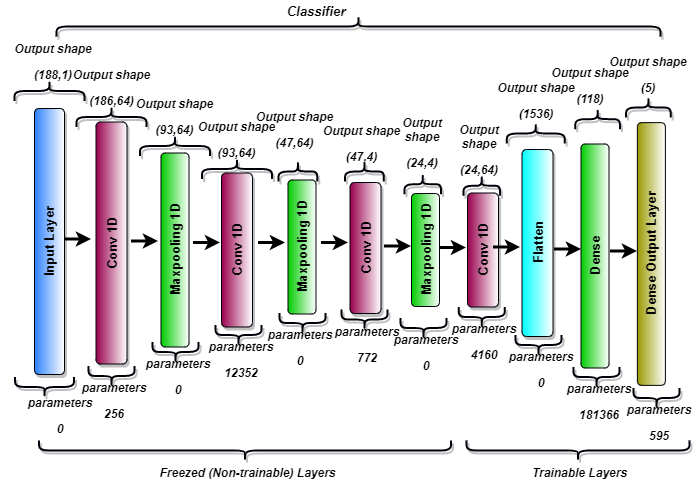}
\caption{The proposed CNN-based classifier}
\label{fig:classifier}
\end{figure*}

The proposed classifier is composed of 4 convolution layers, 3 max pooling layers, 2 fully connected layer and 1 softmax layer for classification, as shown in Figure~\ref{fig:classifier}. The classifier is designed for classifying an input ECG signal into one of the five classes, as shown in Table~\ref{tab:ECG_classes}. We use transfer learning to transfer the encoder part of the trained autoencoder into the proposed classifier, because these convolution layers aim at removing the noise from raw input data and the next layers in the classifier aim to classify the input ECG signal. Hence, the first 3 convolutional layers do not need to be trained while training the individual local classifiers. In other words, we keep the first 3 convolutional layers static during the classifier training phase, which means that no parameters are updated during back propagation in the first 3 convolutional layers. This provides each local node $E_{i}$ with the trained parameters for denoising the signal while training the classifier, which increases the performance of the classier. As for the last 2 convolution layers and the fully connected layers, since they are at a higher level, they focus on learning specific features for the classification task. Therefore, we update their parameters during the classifier training phase. The softmax serves as the classification function, and is given by the following equation:
\begin{equation}
y_{i}=\frac{\exp^{z_{c}}}{\sum_{c=1}^{C}\exp^{z_{c}}},
\end{equation} 
where $C$ is the total number of classes, $z_{c}$ denotes the learned probability for a specific class $c$, and $y_{i}$ is the final classification result for a sample $i$. Our classifier uses categorical cross-entropy (CE) as the loss function. This gives probability over the $C$ classes for each input sample, given by Eq.~\eqref{eq:ce}. Where $t_{c}$ is the ground truth for each class $c$.
\begin{equation}\label{eq:ce}
\text{CE}=-\sum_{c}^{C}t_{c}log(y_{i})
\end{equation}

\begin{table}[!ht]
\centering
\caption{The five classes of ECG signals}
\label{tab:ECG_classes}
\begin{tabular}{r l}
\toprule
Class description & Single-letter symbol\\
\midrule
Non-ecotic beats (normal beat) & N\\
Supraventricular ectopic beats & S\\
Ventricular ectopic beats & V\\
Fusion Beats & F\\
Unknown Beats & Q\\
\bottomrule
\end{tabular}
\end{table}

\subsection{XAI with Grad-CAM}

\begin{figure*}[!ht]
\centering
\includegraphics[width=0.8\linewidth]{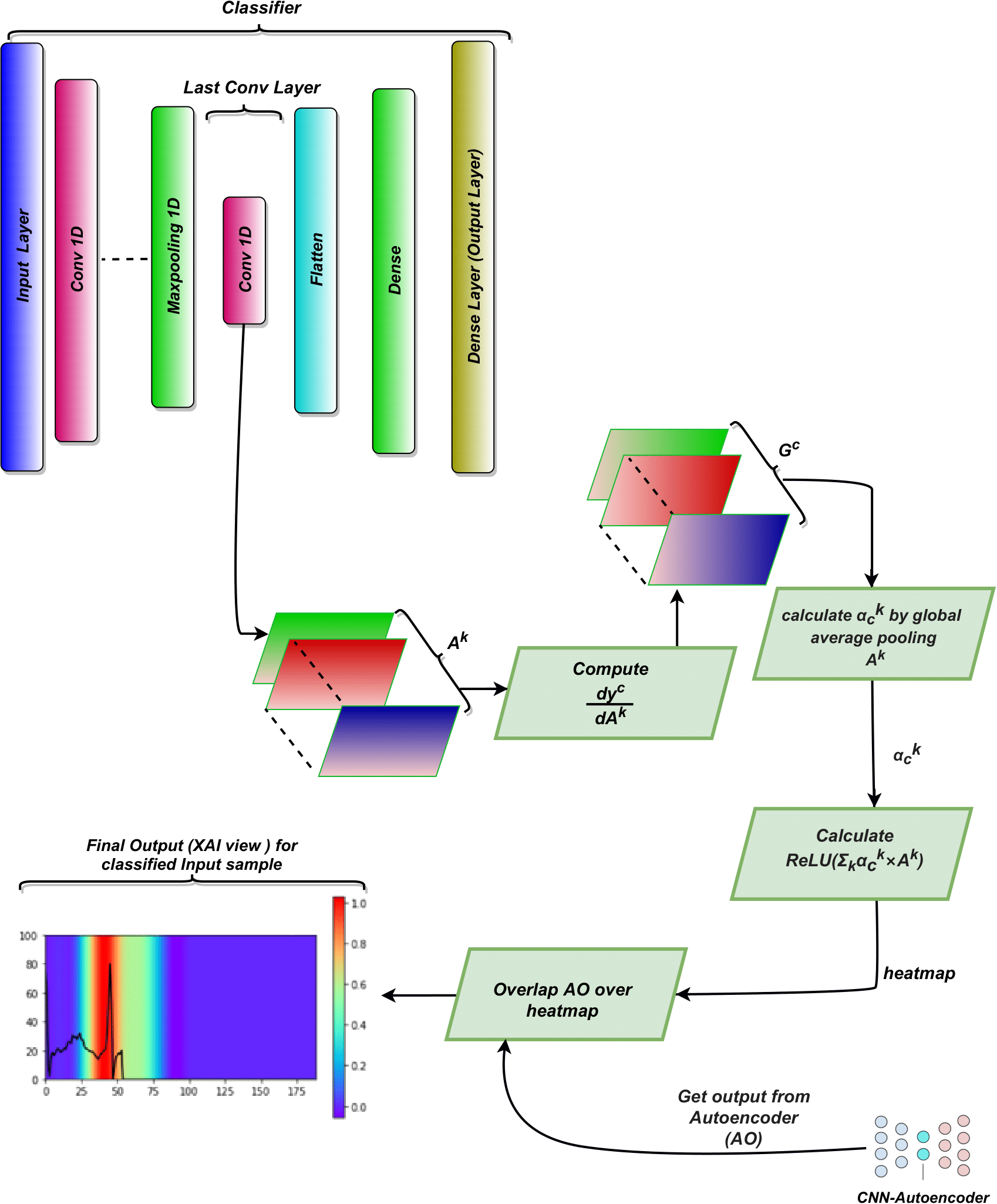}
\caption{Overview of the proposed XAI module in our framework}
\label{fig:gradcam}
\end{figure*}

Inspired by the work in \cite{selvaraju2016grad} and \cite{assaf2019mtex}, we decided to use Gradient-weighted Class Activation Mapping (Grad-CAM) and modified it for time series data on top of our classifier, which uses class-specific gradient information to localize important regions. We combine these localized regions with an existing time-series visualization map to create a high-resolution heatmap visualization. Using this visualization, practitioners can understand the reason of a certain prediction given by the classifier. The XAI with GRAD-CAM module is shown in Figure~\ref{fig:gradcam}.

The creation of this heatmap visualization consists of the following steps:
\begin{enumerate}
\item In the first step, we compute the gradient of $y^{c}$ (where $y^{c}$ is the score for any class $c$) with respect to the feature map activations $A^{k}$ for kernel $k$ of the last convolution layer. If $G_{c}$ represents the gradients for any class $c$, it can be represented as follow:
\begin{equation}
G_{c}= \frac{\partial y^{c}}{\partial A^{k}}.
\end{equation} 
Any particular value calculated in this step depends on the input ECG signal (sample input). The weights of the classifier are fixed at this stage. We first reshape an input sample into the batch size and feed it into the classifier, since the input determines the feature maps $A_{k}$ as well as $y^{c}$.
	
\item The second step consist of global average pooling of the gradients $G_{c}$, both along height $h$ and width $w$ to obtain the neuron importance weights $\alpha_{k}^{c}$ also called alpha values, given by Eq.~\eqref{eq:alpha}.
\begin{equation}\label{eq:alpha}
\alpha_{k}^{c}=\frac{1}{Z}\sum_{h}\sum_{w} \frac{\partial y^{c}}{\partial A^{k}}
\end{equation}
These alpha values for class $c$ and feature map $k$ will be used later as a weight applied to the feature map $A^{k}$.
	
\item The third step consist of weighted linear combination of the feature map activations $A^{k}$ and $\alpha_{k}^{c}$ is calculated using the alpha values, given by Eq.~\eqref{eq:gra}.
\begin{equation}\label{eq:gra}
\text{Grad\_CAM}^{c}=\text{ReLU}(\sum_{k}\alpha_{k}^{c}A^{k})
\end{equation}
This gives us the final Grad-CAM heatmap. A rectifier linear Unit (ReLU) function is applied to emphasize only the positive values and turn all the negative values into 0.
	
\item The classifier's last convolutional layer's features are quite small, and it is difficult to visualize them for analysis. To address this problem, we upsample the heatmap to the size of the input sample in width. Moreover, we feed the input sample to the autoencoder and receive a denoised version of the input sample and overlap it on the heatmap. In the resulting heatmap, regions overlapping between the heatmap and the ECG signal show the point of focus during prediction. This gives a detailed picture to the practitioners to understand which region of the ECG input signal the classifier is looking at while making a prediction.
\end{enumerate}

\subsection{Learning Process}
\label{subsec:Learning_Process}

The learning process of the proposed method has been depicted in Figure~\ref{fig:framework_overview}. For a clearer explanation, we present the learning procedure in Algorithm~\ref{algo:algorithm}. It should be noted that the algorithm works continuously with new emerging data. Optionally, if an $E_{i}$ wants to personalize the classifier $C$, it can be done by keeping all the convolution layers of the final updated classifier static and by training the dense layers for personalization. This is  because the convolution layers aim at extracting low-level features about activity recognition and for the densely connected layers, since they are at a higher level, they focus on learning specific features for the task and the user.

\begin{algorithm}[!ht]
	\caption{Training procedure of Proposed method}
		\label{algo:algorithm}
	\DontPrintSemicolon
	
	\KwInput{Data from edge nodes {$D_{1}, D_{2}, \ldots, D_{n}$}} 
	\KwOutput{Trained aggregated and updated model}
	{Global Server $G_{s}$ constructs the initial Global Autoencoder AE and compiles it using the predefined hyper-parameters} \\
	{ $G_{s}$ waits for the $E_{i}$ to request. If request received, send AE to the $E_{i}$}\\
	{$E_{i}$ receives the AE, and trains it on  its local data $D_{i}$ and sends trained weights of AE back to $G_{s}$ }\\
	$G_{s}$, wait for $n$  $E_{i}$ to send back their locally train AE. 
	
	\If {weights received form $n$ $E_{i}$}
	{
		$F(w)=\sum_{k=1}^{n}\frac{n_{k}}{n_{t}}w_{r+1}^{k}$
	} 
	{$G_{s}$ constructs a classifier $C$}\\
	\For { For $i=1,2,3$ }{ set Weight of convolutional $Layer_{i}$  of $C$ = Weight of convolutional $\text{Layer}_{i}$ of $F(w)$.\\
		set convolutional $\text{Layer}_{i}$ of $C$ trainable = False}
	{$G_{s}$ sends $C$ to $E_{i}$ }\\
	$E_{i}$ trains $C$ on $D_{i}$ and sends back the trained $C$ to $G_{s}$.\\
	$G_{s}$, wait for $n$ $E_{i}$ to send back their locally trained $C$.
	
	\If {weights received form $n$ $E_{i}$}
	{
		$F(w)=\sum_{k=1}^{n}\frac{n_{k}}{n_{t}}w_{r+1}^{k}$
	}
	$G_{s}$ send $F(w)$ to $E_{i}$\\
	$E_{i}$ set $F(w)$ as weight of $C$ and makes predictions \\
	Repeat with continuously emerging data.
\end{algorithm} 

The global server $G_{s}$ (Aggregation Server) creates an autoencoder AE with predefined hyper-parameters. It should be noted that we use Keras auto-tuner to get the best possible hyper-parameters. Keras auto-tuner empirically tries to find the best possible hyper-parameters. After creating the AE, $G_{s}$ waits for the clients' request. When clients request $G_{s}$, it sends the AE to the client. It is worth noting that, each global round is divided into two tiers, for the first tier $G_{s}$ sends the AE and for the second tier $G_{s}$ sends the classifier $C$. Hence while requesting, each client mentions the tier as well. On receiving the autoencoder, client $E_{i}$ trains the autoencoder on its local data $D_{i}$. When the training is completed the client sends back the trained weights of the autoencoder to the global server. The server waits for a fixed number $n$ of clients to send the weights of their locally trained AE. Here, $n$ can be decided by mutual consensus among administrators. When the desired number of clients send their weights and are received by $G_{s}$, it aggregates the weights of all the clients by using the formula for aggregation given by Eq.~\eqref{eq:agg} from \cite{mcmahan2017communication}.
\begin{equation}\label{eq:agg}
F(w)=\sum_{k=1}^{n}\frac{n_{k}}{n_{t}}w_{r+1}^{k},
\text{ where }
F_{k}(w)=\frac{1}{n_{k}}\sum_{i\in P_{k}}f_{i}(w)).
\end{equation} 
Here, $F(w)$ are the aggregated weights, $n_{t}$ is number of data samples of all participants and $n_{k}$ is the number of samples of $k^{\text{th}}$ participant. For a machine learning problem, typically $f_{i}(w) =(x_{i}, y_{i}; w)$, that is, the loss of the prediction on example $x_{i}, y_{i}$ made with model parameters $w$. There are $n$ clients over which the data is partitioned, with $P_{k}$ the set of indexes of data points on client $k$, $n$ is total number of participants in each round and $r$ is the global round number.

After aggregation, $G_{s}$ creates a new CNN-based classifier $C$ for classification. Here, again we use the Keras auto-tuner for best hyper-parameters for the newly created $C$. Furthermore, we use the encoder part of the autoencoder for transfer learning. We transfer the weights of the updated and aggregated encoder part of AE to $C$ and set the transferred layers to static. After this, $G_{s}$ sends $C$ to each client $E_{i}$. Upon receiving $C$, each $E_{i}$ trains the classifier using its local data and sends it back to $G_{s}$. $G_{s}$ collects the weight of $n$ clients and aggregates them using Eq.~\eqref{eq:agg}. After aggregation, it sends the aggregated weights back to each $E_{i}$. Clients set the aggregated weights as new weights of their local $C$, which can be further used for predictions. During predictions, the XAI module taps the gradients and outputs the visual explanation.

\subsection{Communication Cost Reduction and Privacy Enhancement}

In the federated learning setting, training data remain distributed over a large number of clients each with unreliable and relatively slow network connections. For the synchronous protocols in federated learning \cite{smith2017federated}, the total number of communication bits that are required during uplink and downlink communication by each of the $N$ clients during training of the global model is given by:
\begin{equation}
\tau^{\text{up/down}} \in \mathcal{O}(U\times|w|\times(H(\Delta w^{\text{up/down}})+\tau)),
\end{equation}
where $U$ is the total number of updates by each client, $|w|$ is the model size and $H(\Delta w^{\text{up/down}})$ is the entropy of transmitted weights during communication, $\tau$ is the difference between the update size and the minimal update size, given by entropy \cite{sattler2019robust}. Generally, there are three ways to reduce the communication costs: (1) reducing the number of clients $N$, (2) reducing the update size, (3) reducing the number of updates $U$. Hence, research on communication-efficient federated learning can be divided into four groups: model compression, client selection, update reduction, and peer-to-peer learning \cite{xu2021federated}. In order to provide communication-efficient federated learning, we provide a new approach for our proposed architecture called layer selection, which comes under the model compression group. Moreover, layer selection can be added to all of the existing approaches to further reduce the communication costs. The proposed layer selection (communication cost reduction) method is shown in Figure~\ref{fig:optimization}, with more details given below.

Suppose that W1 and W2 represent the weights of all layers of encoder and decoder of the autoencoder, respectively, trained at edge devices. Since we are only concerned with the encoder part of the autoencoder, the edge devices select the weights of the encoder part (W1) and send them to the global server. The global server aggregates the received weights to obtain the global weights, represented as AW1 and sent to the edges. After receiving AW1, the edge devices use transfer learning to transfer these global weights to their local classifier and freeze the transferred layers, as mentioned earlier. The edge devices train the local classifier using their local data. Suppose that WC1 and WC2 represent the weights of the trainable lower (convolutional) and higher (dense) layers of a local classifier, respectively. As the higher layers learn specific features about the underlying data \cite{arpit2017closer}, each edge sends only WC1 to the aggregation server that carries common and low-level features about the training data. The aggregation server performs weighted aggregation of all WC1 weights received to obtain AWC1, which are then sent to edge devices. The edge devices use AWC1 along with their individual WC2 for a more localized classification of the ECG. Since we share few weights compared to the classical method, this makes the communication lighter and reduces the communication costs. Moreover, since the FL framework continuously performs global training with emerging data, our communication cost reduction method can significantly reduce the overall communication costs.

Furthermore, recall that features in deep neural networks are highly transferable in the lower levels of the network since they focus on learning more common and low-level features. As the edge devices only send the weights of lower layers, the privacy of underlying data at each edge is enhanced. To be more precise, the weights of lower layer, weights of the encoder part in the autoencoder (WC1), contains more common and low-level features about the underlying data, while the weights of higher layers, weights of the decoder part of the autoencoder (WC2), contains more specific features about the underlying data. Hence, by not communicating WC2, we can increase the privacy of the local data by sharing only weights (WC1) that contains more common and low-level (i.e., less private) features.

\begin{figure*}[!ht]
\centering
\includegraphics[width=0.8\linewidth]{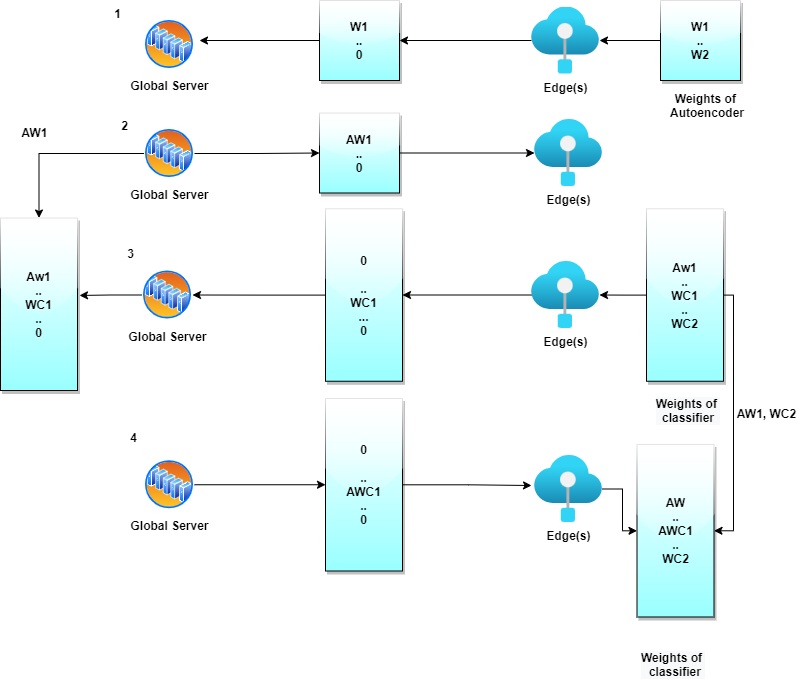}
\caption{The layer selection method for communication cost reduction}
\label{fig:optimization}
\end{figure*}

\section{Experimental Results}
\label{sec:experiments}

\subsection{Dataset}

\begin{figure}[!ht]
\centering
\includegraphics[width=\linewidth]{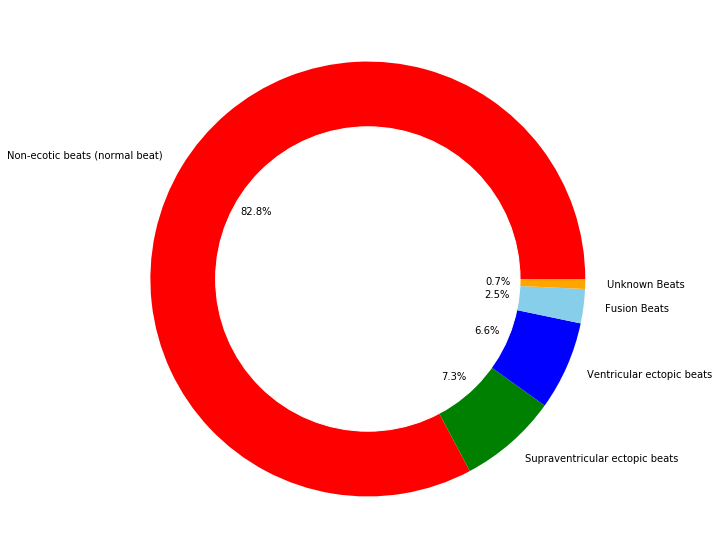}
\caption{The distribution of the original dataset}
\label{fig:unbalanced_data}
\end{figure}

\begin{figure}[!ht]
\centering
\includegraphics[width=\linewidth]{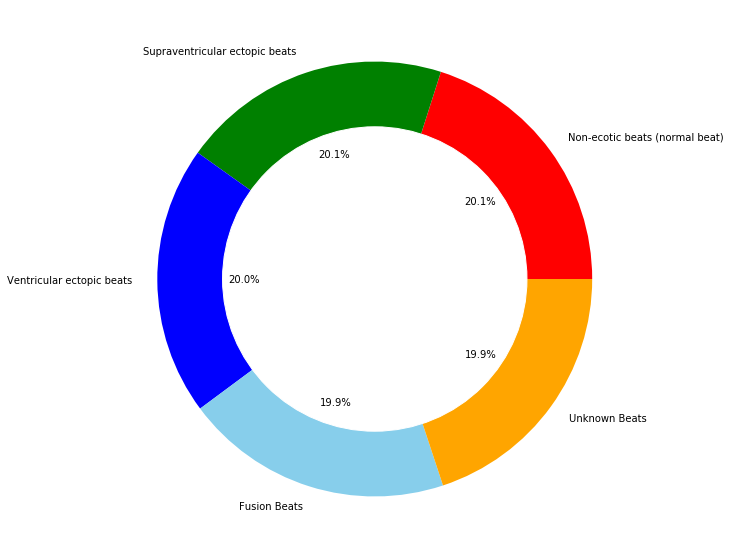}
\caption{The distribution of the upsampled (re-balanced) dataset}
\label{fig:balanced_data}
\end{figure}

\begin{figure*}[!ht]
\centering
\includegraphics[width=\linewidth]{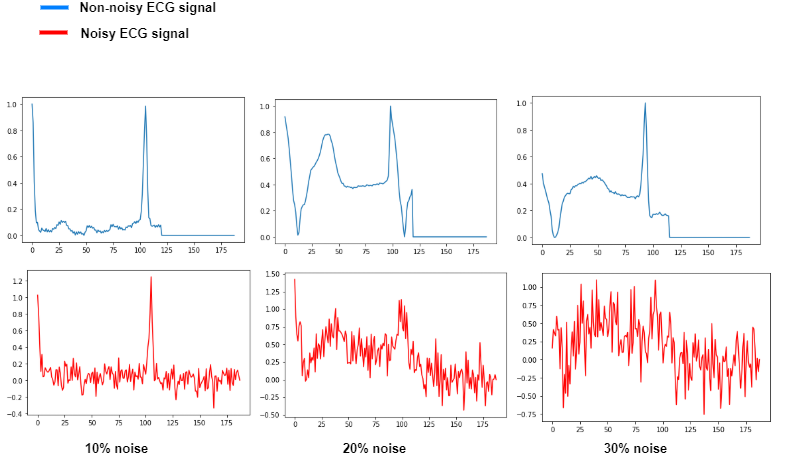}
\caption{Comparison of the original and the noisy version of the dataset}
\label{fig:signals}
\end{figure*}

For the experimental purpose, we used the widely used MIT-BIH Arrhythmia Database \cite{moody2001impact} as our baseline dataset. This database contains 48 half-hour excerpts of two-channel ambulatory ECG recordings, obtained from 47 subjects studied by the BIH Arrhythmia Laboratory between 1975 and 1979. The dataset includes 109,446 samples. Twenty-three recordings were chosen at random from a set of 4,000 24-hour ambulatory ECG recordings collected from a mixed population of inpatients (about 60\%) and outpatients (about 40\%) at Boston's Beth Israel Hospital; the remaining 25 recordings were selected from the same set to include less common but clinically significant arrhythmias that would not be well-represented in a small random sample. In our experiment, we have used ECG lead II re-sampled to the sampling frequency of 125 Hz as the input. It should be noted that this dataset has unbalanced classes. Figure~\ref{fig:unbalanced_data} shows the distribution of the original dataset. This highly unbalanced data can cause problems like overfitting. Hence to balance the classes we used upsampling. The resulting data distribution after upsampling is shown in Figure~\ref{fig:balanced_data}. Furthermore, this dataset is highly preprocessed, but in real-world scenarios, the ECG data collected is always noisy. Hence, to simulate more realistic data we introduced 10-30\% noise into the original dataset and trained the proposed framework on the noisy version of the dataset, too. A comparison of the original (clean) and noisy datasets is shown in Figure~\ref{fig:signals}.

\subsection{Implementation Details}

The Framework was implemented using Python and TensorFlow. Secure socket layer communication was used for communication between the server and edge devices. Both the autoencoder and the classifier were trained locally only on three local Raspberry Pi devices (Pi 3 Model B+ with 1.4GHz, 64-bit quad-core ARMv8 CPU and 1GB LPDDR2 SDRAM), denoted by $\text{Edge}_{1}$, $\text{Edge}_{2}$ and $\text{Edge}_{3}$. Furthermore, a workstation with an Intel core i-6700HQ CPU and 32 GB RAM was used as the global server $G_{s}$. It should be noted that FedHealth \cite{chen2020fedhealth} initially trained their model at $G_{s}$, which may cause security risks in the case of a malicious global server. If the models (AE and $C$) are trained initially on $G_{s}$ this may cause biased training. Hence, to avoid such risks, we performed only aggregation at the $G_{s}$. Furthermore, AE adopted a convolution size of 3. It uses a Root Mean Square Propagation (RMSProp) as the optimizer. Each $E_{i}$ device uses 80\% of data for training and 20\% of data for evaluation. We distributed the dataset randomly at each edge device and introduced random noise. In this case, the data in $\text{Edge}_{1}$ contains 20\% random noise, the data in $\text{Edge}_{2}$ contains 30\% random noise,the data in $\text{Edge}_{3}$ contains 10\% random noise. Furthermore, each edge used a fixed batch size of 100, and was trained for 50 training epochs. Moreover, each edge used an evolving learning rate, given by Eq.~\eqref{eq:lr}.

The classifier $C$ used a batch size of 100. The learning rate was set to 0.001 with 150 training epochs. The accuracy of each of the locally trained $C$ was calculated by using the following equation:
\begin{equation}
A^{i}_{cc}=\frac{|x:x\in D_{i}\wedge y^{'}(x)=y(x) |}{|x:x\in D_{i}|}.
\end{equation}
In regards to the execution time, given the above setting, it took an average of 745 seconds to complete one global round of training. Furthermore, it took an average of 2.32 seconds to generate prediction and XAI results. \color{black}
\section{Performance Analysis of the Proposed Method}
\label{sec:Analysis}

In this section we analyze performance of the proposed framework using some state-of-the-art metrics.

\subsection{Reconstruction of Autoencoder}

\begin{figure*}[!ht]
\centering
\includegraphics[width=0.9\linewidth]{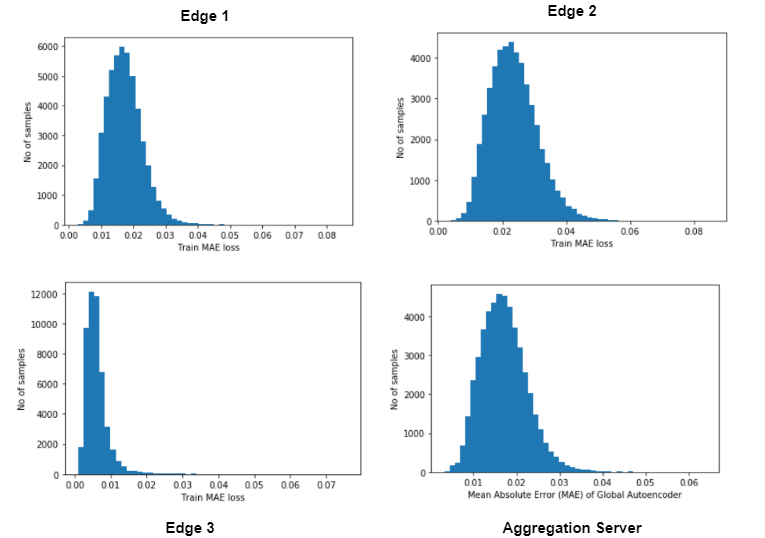}
\caption{Reconstruction MAE}
\label{fig:reconstruction_MAE}
\end{figure*}

We introduced noise in to the dataset and used the noisy sample as the input in the autoencoder and the cleaned samples as labels. The performance of the autoencoder was measured using reconstruction mean absolute error (MAE). Reconstruction MAE for each locally trained AE in each of $\text{Edge}_{1}$, $\text{Edge}_{2}$, $\text{Edge}_{3}$ and aggregated AE is given in Figure~\ref{fig:reconstruction_MAE}. It can be seen that the reconstruction MAE of the aggregated autoencoder is nearly 0, which means that our autoencoder reconstructed the original signal very well. Moreover, it can be seen that reconstruction MAE aggregation AE is less than or nearly equal to the reconstruction MAE of each locally trained AE.

\subsection{Classification Performance}

Classification performance was measured using the four standard metrics found in the literature \cite{hu1997patient}: classification accuracy, precision, recall and F1-score. While accuracy measures the overall system performance over all classes in the dataset, the other metrics are specific to each class, and they measure the ability of the classification algorithm to distinguish certain events. For a binary classifier, each of the metrics is defined as follows:
\begin{enumerate}
\item \textbf{Accuracy} is the most intuitive performance measure and it is simply a ratio of correctly predicted observation to the total observations, as defined below:
\begin{equation}
\text{Accuracy} = \frac{\text{TP}+\text{TN}}{\text{TP}+\text{FP}+\text{FN}+\text{TN}},
\end{equation}
where TP, TN, FP and FN refer to the numbers of true positives, true negatives, false positives, and false negatives, respectively.
      
\item \textbf{Precision} is the ratio of correctly predicted positive observations to the total predicted positive observations, as defined below:
\begin{equation}
\text{Precision} = \frac{\text{TP}}{\text{TP}+\text{FP}}.
\end{equation}

\item \textbf{Recall} is the ratio of correctly predicted positive observations to all the observations in actual positive class, as defined below:
\begin{equation}
\text{Recall} = \frac{\text{TP}}{\text{TP}+\text{FN}}.
\end{equation}

\item \textbf{F1-score} is the harmonic mean of precision and recall, as defined below:
\begin{equation}
\text{F1-Score} = \frac{2 \times \text{Recall} \times \text{Precision}}{\text{Recall} + \text{Precision}}.
\end{equation}
\end{enumerate}
The above definitions can be easily extended to multi-class classifiers with $n>2$ classes. For instance, accuracy is defined as the ratio between the number of total correct predictions and the total number of samples. For other metrics, i.e., precision, recall, and F1-score, we can derive $n$ binary classifiers, one for each given class versus all remaining classes (one-vs-rest), and then use the above definitions of the three metrics as usual for each of the $n$ binary classifiers.

Precision, recall, F1-score metrics of each binary classifier (one for each of five class labels) at the three edge devices ($\text{Edge}_1$, $\text{Edge}_2$, and $\text{Edge}_3$) and the global server are given in Table~\ref{tab:overall_performance}. We also show the accuracy of the five-class classifier in the last column. It should be noted that the results shown in Table~\ref{tab:overall_performance} are computed using the noisy data which we prepared earlier. We also tested the proposed classifier using the original (clean) data. With this data it provided $98\pm0.9$\% accuracy. Other metrics, such as precision, recall and F1-score, are shown in Table~\ref{tab:overallresult_clean}. However, for real-time use we expect the data to be noisy, which is why we proceeded with the noisy data.

\begin{table*}
\centering
\captionsetup{width=\linewidth}
\caption{The classification performance of the proposed framework, with the noisy version of the dataset}
\label{tab:overall_performance}

\begin{subtable}[t]{0.45\linewidth}
\centering
\begin{tabular}{|*{5}{c|}}
\hline
\textit{Class} & \textit{Precision} & \textit{Recall} & \textit{F1-Score} & \textit{Accuracy}\\
\hline
N & 89\% & 91\% & 90\% &  \multirow{5}{*}{94.9\%}\\
\cline{1-4}
S & 94\% & 89\% & 92\% &\\
\cline{1-4}
V & 93\% & 96\% & 94\% &\\
\cline{1-4}
F & 95\% & 94\% & 95\% &\\
\cline{1-4}
Q & 99\% & 99\% & 99\% &\\
\hline
\end{tabular}
\subcaption{Edge 1 (20\% noise)}
\end{subtable}
\hfill
\begin{subtable}[t]{0.45\linewidth}
\centering
\begin{tabular}{|*{5}{c|}}
\hline
\textit{Class} & \textit{Precision} & \textit{Recall} & \textit{F1-Score} & \textit{Accuracy}\\
\hline
N & 85\% & 87\% & 86\% &  \multirow{5}{*}{91.9\%}\\
\cline{1-4}
S & 91\% & 87\% & 88\% &\\
\cline{1-4}
V & 91\% & 94\% & 92\% &\\
\cline{1-4}
F & 93\% & 93\% & 93\% &\\
\cline{1-4}
Q & 98\% & 98\% & 98\% &\\
\hline
\end{tabular}
\subcaption{Edge 2 (30\% noise)}
\end{subtable}

\vspace{1em}

\begin{subtable}[t]{0.45\linewidth}
\centering
\begin{tabular}{|*{5}{c|}}
\hline
\textit{Class} & \textit{Precision} & \textit{Recall} & \textit{F1-Score} & \textit{Accuracy}\\
\hline
N & 94\% & 98\% & 96\% &  \multirow{5}{*}{97.9\%}\\
\cline{1-4}
S & 98\% & 92\% & 95\% &\\
\cline{1-4}
V & 95\% & 99\% & 97\% &\\
\cline{1-4}
F & 99\% & 94\% & 96\% &\\
\cline{1-4}
Q & 97\% & 100\% & 98\% &\\
\hline
\end{tabular}
\subcaption{Edge 3 (10\% noise)}
\end{subtable}
\hfill
\begin{subtable}[t]{0.45\linewidth}
\centering
\begin{tabular}{|*{5}{c|}}
\hline
\textit{Class} & \textit{Precision} & \textit{Recall} & \textit{F1-Score} & \textit{Accuracy}\\
\hline
N & 90\% & 92\% & 91\% &  \multirow{5}{*}{94.5\%}\\
\cline{1-4}
S & 94\% & 89\% & 91\% &\\
\cline{1-4}
V & 93\% & 96\% & 94\% &\\
\cline{1-4}
F & 96\% & 96\% & 95\% &\\
\cline{1-4}
Q & 99\% & 99\% & 99\% &\\
\hline
\end{tabular}
\subcaption{Global/Aggregation Server }\label{tab:server}
\end{subtable}
\end{table*}

\begin{table}[!htb]
\centering
\caption{The classification performance of the proposed framework, with the original (clean) dataset}
\label{tab:overallresult_clean}
\begin{tabular}{|*{5}{c|}}
\hline
\textit{Class} & \textit{Precision} & \textit{Recall} & \textit{F1-Score} & \textit{Accuracy}\\
\hline
N & 95\% & 99\% & 97\% & \multirow{5}{*}{98.9\%}\\
\cline{1-4}
S & 98\% & 97\% & 98\% &\\
\cline{1-4}
V & 97\% & 99\% & 98\% &\\
\cline{1-4}
F & 99\% & 93\% & 96\% &\\
\cline{1-4}
Q & 100\% & 100\% & 100\% &\\
\hline
\end{tabular}
\end{table}

\subsection{Qualitative Analysis}

\begin{figure*}[!ht]
\centering
\includegraphics[width=\linewidth]{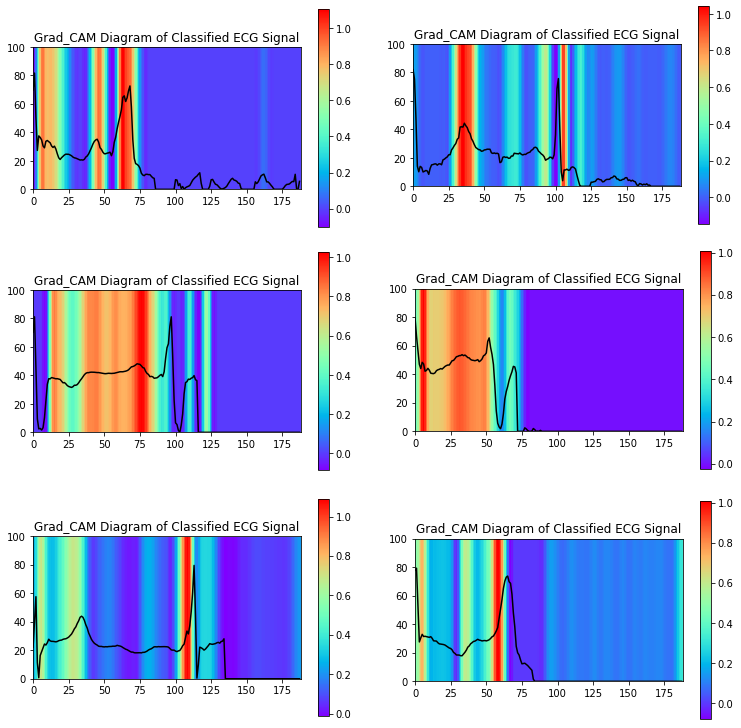}
\caption{The outputs of the XAI module}
\label{fig:output_grad_cam}
\end{figure*}

Understanding the reasons for predictions of the model decision is very important in healthcare applications. In order to validate that the decisions made by the proposed XAI module are interpretable, we use visualization to demonstrate that clinically important beats in the ECG wave are used for classification. Figure~\ref{fig:output_grad_cam} illustrates the importance for each beat that the ECG classifier is giving while performing classification of some instance ECG signal inputs.

\begin{figure}[!ht]
\centering
\includegraphics[width=\linewidth]{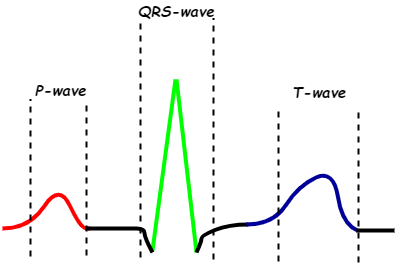}
\caption{The major waves of a single normal ECG pattern}
\label{fig:ecg_signal}
\end{figure}

In order to achieve the interpretability/explainability of the XAI module, it is important to understand the ECG signal \cite{conover2002understanding}. Generally, the amplitude and width of the p-wave, QRS complex and the T-wave are important features of an ECG graph, as shown in Figure~\ref{fig:ecg_signal}. These regions play a vital role in ECG analysis \cite{berkaya2018survey}. The XAI module in the proposed framework shows that the proposed classifier looks at these critical features of the input sample. In Figure~\ref{fig:output_grad_cam} , the red segments show more important regions of the heartbeat for the network's decision while predicting a particular class. In other words, the red segments of the heartbeat have more influence on the detection process of the classifier while classifying the input ECG signal.\color{black} These results can be used to help clinical practitioners to diagnose the underlying health issues. However, we strongly advise that these results should not be used for any medical consultation without prior discussion with a clinical professional. In other words, heat maps should be cross-checked by clinicians with prior expert knowledge.

\subsection{Comparison With Other State-of-the-Art Methods}

We compared our proposed framework with the state-of-the-art methods reported in 2020 \cite{chen2020fedhealth,oh2018automated, liaqat2020detection,erdenebayar2019automatic,atal2020arrhythmia,acharya2017automated,yao2017atrial,mousavi2020han,nurmaini2020robust}. First we compare the previous work with ours to show that the proposed framework provides all of the desirable properties like interpretability, privacy preserving, and working with raw data. Table~\ref{tab:comparision} shows the comparison between our proposed method and the other methods. It can be seen that the proposed scheme outperforms others by providing all desirable properties, while others lack some of the desirable properties. Moreover, we also compare our work with existing works for ECG classification. It should be noted that other methods used the baseline MITBIH dataset (without noise), with which better accuracy results can be achieved. Contrastingly, we introduced (10\%-30\%) noise into the data to make it more realistic. Table~\ref{tab:comparisionwithecg} shows the comparison of classification performance between our proposed method and the other methods. It can be seen that, our proposed method outperformed the methods in others. It should be noted that the proposed classifier deals with the classification tasks of five classes, while others deal with fewer classes (some with two and some with three). Additionally, the proposed method works in federated architecture and performs better compared to others.  Our proposed method provides explainability as an additional feature. Moreover, the proposed method provides data privacy to the users via the federated setting, which is not the case for other methods. Furthermore, the proposed method can denoise raw signals without any preprocessing, followed by classification and explainability.

\begin{table}[!ht]
\centering
\caption{Comparison with the state-of-the-art work}
\label{tab:comparision}
\begin{tabular}{|c|c|c|>{\centering\arraybackslash}m{0.2\linewidth}|}
\hline 
Scheme & Interpretability & Raw Input & Privacy Preserving  \\ \hline 
\cite{pyakillya2017deep} & \xmark & \xmark & \xmark\\ \hline
 \cite{mathews2018novel} & \xmark & \xmark & \xmark\\ \hline
 \cite{yuan2016automated} & \xmark & \xmark & \xmark\\ \hline
\cite{chen2020fedhealth} & \xmark & \xmark & \cmark\\ \hline
\cite{oh2018automated} & \xmark & \xmark & \xmark\\ \hline
\cite{liaqat2020detection} & \xmark & \xmark & \xmark\\ \hline
 \cite{erdenebayar2019automatic} & \xmark & \xmark & \xmark\\ \hline
\cite{atal2020arrhythmia} & \xmark & \xmark & \xmark\\ \hline
 \cite{acharya2017automated} & \xmark & \xmark & \xmark\\ \hline
\cite{yao2017atrial}& \xmark & \xmark & \xmark\\ \hline
\cite{mousavi2020han} & \cmark & \xmark & \xmark\\ \hline
\cite{nurmaini2020robust} & \xmark & \xmark & \xmark\\ \hline
Proposed & \cmark & \cmark & \cmark\\ \hline
\end{tabular}
\end{table}

\subsection{Privacy Enhancement}

As mentioned earlier, digital healthcare data is like digital finger prints that carry a lot of personal information. Hence, we should protect such data as much as possible while using them in machine learning algorithms. Most past studies on ECG classification do not provide privacy protection of such data because they are centralized and data are shared with the central model directly. Recently, Chen et al.~\cite{chen2020fedhealth} used federated learning to provide privacy protection, by only sharing the learned parameters without sharing the data. Although the shared parameters can protect privacy, there are still chances to recover some information from the shared parameters of higher level layers in the classifier, since they can contain more data-specific information as discussed previously. As a comparison, in our proposed framework we only share the learned parameters from lower-level layers that carry only more common and low-level (i.e., less privacy-sensitive) features. Thus, our proposed framework can enhance privacy even further, and at the same time can reduce communication costs as fewer parameters are shared between the edge/local and global servers.

A comparison between with existing work in federated setting for healthcare is shown in Table~\ref{tab:security}.

\begin{table}[!ht]
\centering 
\caption{Comparison with previous studies for ECG classification}
\label{tab:comparisionwithecg}
\begin{tabular}{|c|*{3}{>{\centering\arraybackslash}m{0.21\linewidth}|}}
\hline  
Scheme & Centralized or Federated & Acc (clean data) & Acc (noisy data)\\ \hline  
\cite{pyakillya2017deep} & Centralized & 86.0\% & -\\ \hline
\cite{mathews2018novel} & Centralized & 96.9\% & -\\ \hline
\cite{oh2018automated} & Centralized & 98.1\% & -\\ \hline
\cite{liaqat2020detection} & Centralized & 96.5\% & -\\ \hline
\cite{atal2020arrhythmia} & Centralized & 93.1\% & -\\ \hline
\cite{erdenebayar2019automatic} & Centralized & 98.7\% & -\\ \hline
\cite{acharya2017automated} & Centralized & 94.9\% & -\\ \hline
\cite{yao2017atrial} & Centralized & 98.1\% & -\\ \hline
\cite{xia2018detecting} & Centralized & 98.6\% & -\\ \hline
\cite{yuan2016automated} & Centralized & 98.3\% & -\\ \hline
\cite{mousavi2020han} & Centralized & 98.8\% & -\\ \hline
Proposed & Federated & 98.9\% & 94.5\% \\\hline
\end{tabular}
\end{table}

\subsection{Communication Cost Reduction}

Here, we show the communication cost reduction using the proposed communication cost reduction method. The number of total parameters communicated between an edge device and the global server, for one global round, denoted by \text{TPC}, is given as follow:

\begin{multline}
\text{TPC} = \text{W1}+\text{W2} +\text{WC1}+\text{WC2} \\
+\text {AW1}+\text{AW2} +\text {AWC1}+\text{AWC2},
\end{multline}

In the proposed framework, the TPC is given as follow:
\begin{multline}\label{eq:TPC_without_layer_selection}
\text{TPC} = 13386+13429+4160+181961+\\
13386+13429+4160+181961 = 425872,
\end{multline}

With the proposed communication cost reduction method, the TPC is given as follow:
\begin{equation}
\text{TPC} = \text {W1}+ +\text {WC1}+\text {AW1}+ +\text {AWC1},
\end{equation}
In the proposed framework, the TPC is given as follow:
\begin{equation}\label{eq:TPC_with_layer_selection}
\text{TPC} = 13386 + 4160 + 13386 + 4160 = 35092,
\end{equation}
From Eqs.~\eqref{eq:TPC_without_layer_selection} and \eqref{eq:TPC_with_layer_selection}, we can calculate that the proposed communication cost reduction method reduces the communication cost by 8.2\%.  

\begin{table}[!ht]
\centering
\caption{Comparison with the state-of-the-art work in federated setting for healthcare}
\label{tab:security}
\begin{tabular}{|c|>{\centering\arraybackslash}m{0.35\linewidth}|>{\centering\arraybackslash}m{0.3\linewidth}|}
\hline 
Scheme & 
Communication Cost Reduction & Privacy Enhancement\\\hline
\cite{chen2020fedhealth} & 
\xmark & \xmark\\\hline
Proposed & 
\cmark & \cmark\\\hline
\end{tabular}
\end{table}

\subsection{Time Complexity of proposed Algorithm}
In this section, we provide the time complexity of the proposed Algorithm \ref{algo:algorithm}. In a CNN-based network, the number of features in each feature map is at most a constant times the number of input features let us say $n$ (typically the constant is < 1). Convolving a fixed-size filter across an input signal with $n$ features takes $O(n)$ time, since each output is just the sum product between some features let’s say $k$ in the input, and a fixed number of weights $w$ in the filter, and $w$ and $k$ do not vary with $n$. Similarly, any max or average pooling operation does not take more than a linear amount of time in the input size. Moreover, the edge node can compute in parallel, therefore, the overall runtime is still linear i.e., $O(n)$.
\subsection{Limitations and Future work}
In this section, we discuss some limitations of the proposed framework. Our proposed framework provides a unique way of diagnosing arrhythmias with the explanation of the predicted results. However, there are certain limitations, which can be explored in the future to make the proposed framework more reliable. First, in federated learning, data is not collected at a single server and there are multiple devices for collecting and analyzing data. Such distributed settings increases the chances of data poisoning attacks, hence, methods should be developed for data integrity and authentication. Second, the proposed framework considers the data and devices in distributed edges to be homogeneous, but in some cases, the data and devices may be heterogeneous, and device-specific characteristics may limit the generalizability of the local models from device to device and may reduce the accuracy of the aggregated model. Furthermore, from a security perspective, malicious internal attackers have been not considered in our work, who may attack other peers by inserting hidden back-doors into the joint global model. Last but not the least, in case the global server goes down, the whole system may stop working. As part of our future work, we aim to improve the current framework by considering the above-mentioned limitations to produce a more secure and robust framework.        
\color{black}
\section{Conclusions}
\label{sec:conclusion}

In this paper, we proposed a privacy-preserving, efficient and interpretable/explainable AI-based end-to-end framework to address the limitations of deep learning applications for ECG signal classification. Firstly, we proposed a CNN-based autoencoder in a federated architecture to denoise the raw ECG signal from patients. When trained on the baseline dataset, The proposed autoencoder provided an excellent reconstruction of the raw input signals and improved the overall performance when applied in federated settings. Secondly, we proposed a new classifier for ECG signals. When the classifier was trained in federated settings it was able to improve the overall classification performance of the edge devices. Moreover, the experimental results on the baseline database revealed that the proposed framework achieved outperformed existing algorithms, including both centralized and federated ones. Furthermore, we extended the usability of our framework by providing a novel explainable module on top of the classifier, whose usefulness is visually demonstrated by showing that clinically meaningful heartbeat segments of ECG signals are indeed behind the classification results. Additionally, we also proposed a communication cost reduction method, which can significantly reduce communication costs while increasing the level of privacy protection of users' ECG data against the global server. Hence, the proposed framework shows its applicability by providing many desirable properties including interpretability, privacy protection, communication cost reduction, and high accuracy in classification. Such a combination of such properties does not hold for other existing solutions, therefore making the proposed framework a unique solution for real-world healthcare applications where ECG signal classification is an important task.

Eventually, the proposed framework will encourage (1) more healthcare data owners to participate in training a good machine learning model for patients and health professionals, with fewer privacy concerns, (2) more accurate diagnostic assistance in places with scarce access to cardiologists or healthcare facilities, (3) more interpretable classification results that can be used to identify new potential patterns leading to trigger heart arrhythmias. Hence, the proposed framework has great potential to be added to hospital computer software platforms to support the work of health professionals and ultimately reduce mortality and save human lives.

As a future research direction, we aim at applying the proposed framework to more healthcare applications, especially human activity recognition and anomaly detection in the context of home care, and other types of arrhythmia to extract new patterns that might be helpful for their diagnosis and monitoring. We also aim at extending the applicability of the proposed work by addressing the limitations of the proposed framework as mentioned previously. 

\section*{Acknowledgments}

This research work was supported by the I-SITE Université Lille Nord-Europe 2021 of France under grant No.~I-COTKEN-20-001-TRAN-RAZA.


\bibliographystyle{elsarticle-num}
\bibliography{main}

\end{document}